\documentclass[11pt,a4paper]{article}
\usepackage[hyperref]{emnlp-ijcnlp-2019}
\usepackage{times}
\usepackage{latexsym}
\usepackage{url}
\usepackage{graphicx}
\usepackage{amsmath}
\usepackage{amssymb}
\usepackage{subfigure}
\usepackage{caption}
\usepackage{algorithm}
\usepackage{algorithmicx}
\usepackage{algpseudocode}
\usepackage{multicol}
\usepackage{multirow}
\usepackage{booktabs}
\usepackage{listings}
\usepackage{array}
\usepackage{bm}

\aclfinalcopy 

\title{Adapting Meta Knowledge Graph Information for \\ Multi-Hop Reasoning over Few-Shot Relations}

  \newcommand*{\email}[1]{\texttt{#1}}

  \author{
    \textbf{Xin Lv}$^{1,2,3}$, \textbf{Yuxian Gu}$^{1}$, \textbf{Xu Han}$^{1,2}$, \textbf{Lei Hou}$^{1,2,3}$\thanks{\quad Corresponding Author}\hspace{0.5em}, \textbf{Juanzi Li}$^{1,2,3}$, \textbf{Zhiyuan Liu}$^{1,2}$\\
    $^1$Department of Computer Science and Technology, Tsinghua University \\
    $^2$KIRC, Institute for Artificial Intelligence, Tsinghua University \\
     $^3$Beijing National Research Center for Information Science and Technology \\
    \email{\{lv-x18,gu-yx17,hanxu17\}@mails.tsinghua.edu.cn}\\
    \email{\{houlei,lijuanzi,liuzy\}@tsinghua.edu.cn}
    }

\date{}

\begin{document}
\maketitle
\begin{abstract}
Multi-hop knowledge graph (KG) reasoning is an effective and explainable method for predicting the target entity via reasoning paths in query answering (QA) task. Most previous methods assume that every relation in KGs has enough training triples, regardless of those few-shot relations which cannot provide sufficient triples for training robust reasoning models. In fact, the performance of existing multi-hop reasoning methods drops significantly on few-shot relations. In this paper, we propose a meta-based multi-hop reasoning method (Meta-KGR), which adopts meta-learning to learn effective meta parameters from high-frequency relations that could quickly adapt to few-shot relations. We evaluate Meta-KGR on two public datasets sampled from Freebase and NELL, and the experimental results show that Meta-KGR outperforms the current state-of-the-art methods in few-shot scenarios. Our code and datasets can be obtained from  \url{https://github.com/THU-KEG/MetaKGR}.

\end{abstract}

\section{Introduction}

Recently, large-scale knowledge graphs (KGs) have been demonstrated to be beneficial for many NLP tasks like query answering (QA). A triple query for QA is generally formalized as $(e_s, r_q, ?)$, where $e_s$ is the source entity and $r_q$ is the query relation. 
For example, given a language query ``\textit{What is the nationality of Mark Twain?}", we can transform it into $(\emph{Mark Twain}, \textit{nationality}, ?)$ and then search the target entity \emph{America} from KGs as the answer.
However, as many KGs are constructed automatically and face serious incompleteness problems~\cite{TransE}, it is often hard to directly get target entities for queries.

To alleviate this issue, some knowledge graph embedding methods~\cite{TransE,ConvE} have been proposed to embed entities and relations into semantic spaces to capture inner connections, and then use the learned embeddings for final predictions. Although these embedding-based methods have shown strong abilities in predicting target entities for queries, they only give answers and lack interpretability for their decisions~\cite{MultiHop}. In order to make models more intuitive, ~\citet{MINERVA} and~\citet{MultiHop} propose multi-hop reasoning methods, which leverage the symbolic compositionality of relations in KGs to achieve explainable reasoning results. For example, when queried with $(\emph{Mark Twain}, \textit{nationality}, ?)$, multi-hop reasoning models can give not only the target entity \textit{America} but also multi-hop explainable paths $(\emph{Mark Twain}, \textit{bornIn}, \emph{Florida}) \wedge (\emph{Florida}, \textit{locatedIn}, \emph{America})$ as well.

\begin{figure}[t]
\centering
\setlength{\abovecaptionskip}{2pt}
\setlength{\belowcaptionskip}{0pt}
\includegraphics[width=75mm]{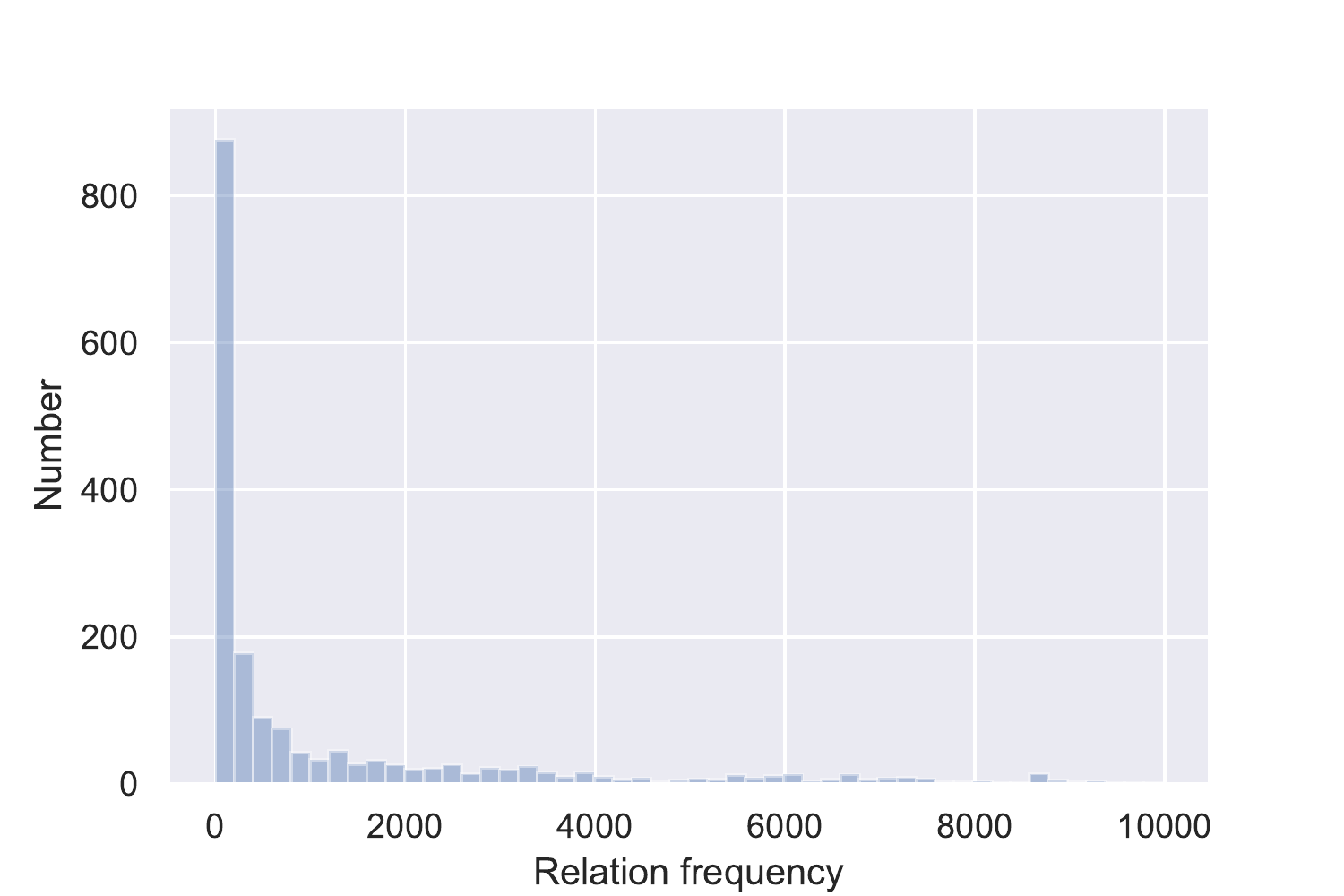}
\caption{The histogram of relation frequency in the real-world knowledge graph Wikidata.}
\label{relation_frequency}
\end{figure}

Most previous work assumes that there are enough triples to train an effective and robust reasoning models for each relation in KGs. However, as shown in Figure \ref{relation_frequency}, a large portion of KG relations are actually long-tail~\cite{one-shot,Fewrel} and only contain few triples, which can be called few-shot relations. Some pilot experiments show that the performance of previous multi-hop reasoning models, e.g., MINERVA~\cite{MINERVA} and MultiHop~\cite{MultiHop}, on these few-shot relations will drop significantly. Note that, there are some knowledge graph embedding models~\cite{one-shot, shi2018open, DKRL} that can deal with zero-shot or one-shot relations, but they still have two main drawbacks: (1) they are embedding-based models and lack interpretability; (2) they focus on zero-shot or one-shot relations, which is a little far away from real-world scenes. In fact, even for people, it is hard to grasp new knowledge with almost no examples. Therefore, few-shot multi-hop reasoning is a quite important and practical problem that has not yet been fully resolved.

In this paper, we propose a meta-based algorithm for multi-hop reasoning (Meta-KGR) to address the above problems, which is explainable and effective for few-shot relations. Specifically, in Meta-KGR, we regard triple queries with the same relation $r$ in KGs as a task. For each task, we adopt reinforcement learning (RL) to train an agent to search target entities and reasoning paths. Similar to previous meta-learning method MAML~\cite{MAML}, we use tasks of high-frequency relations to capture meta information, which includes common features among different tasks. Then, the meta information can be rapidly adapted to the tasks of few-shot relations, by providing a good starting point to train their specific reasoning agents. In experiments, we evaluate Meta-KGR on two datasets sampled from Freebase and NELL, and the experimental results show that Meta-KGR outperforms  the current state-of-the-art multi-hop reasoning methods in few-shot scenarios.

\section{Problem Formulation}
\label{problem}

We formally define a KG $\mathcal{G} = \{\mathcal{E}, \mathcal{R}, \mathcal{F}\}$, where $\mathcal{E}$ and $\mathcal{R}$ denote the entity and relation sets respectively. $\mathcal{F} = \{(e_s, r, e_o)\} \subseteq \mathcal{E} \times \mathcal{R} \times \mathcal{E}$ is the triple set, where $e_s, e_o \in \mathcal{E}$ are entities and $r \in \mathcal{R}$ is a relation between them. In this paper, if the number of triples mentioning a relation $r$ is smaller than the specific threshold $K$, $r$ is considered as a few-shot relation, otherwise, it is a normal relation.

Given a query $(e_s, r_q, ?)$, where $e_s$ is the source entity and $r_q$ is the few-shot relation, the goal of few-shot multi-hop reasoning is to predict the right entity $e_o$ for this query. Different from the previous knowledge graph embedding work, multi-hop reasoning also gives a reasoning path from $e_s$ to $e_o$ over $\mathcal{G}$ to illustrate the whole reasoning process.

\section{Related Work}

\textbf{Knowledge Graph Embedding}~\cite{TransE,DistMult,ConvE} aims to embed entities and relations into low-dimensional spaces, and then uses embeddings to define a score function $f(e_s, r, e_o)$ to evaluate the probability that a triple is true. Recently, several models~\cite{DKRL,shi2018open} incorporate additional entity descriptions to learn embeddings for unseen entities, which can be seen as zero-shot scenarios. \citet{one-shot} predict new facts under a challenging setting where only one training triple for a given relation $r$ is available, which can be seen as a one-shot scenario. Although these models are effective, they lack interpretability for their decisions.

\textbf{Multi-Hop Reasoning} over KGs aims to learn symbolic inference rules from relational paths in $\mathcal{G}$ and has been formulated as sequential decision problems in recent years. DeepPath \cite{DeepPath} first applies RL to search reasoning paths in KGs for a given query, which inspires much later work (e.g., MINERVA~\cite{MINERVA} and DIVA \cite{chen2018variational}). Because it is hard to train an RL model, Reinforce-Walk~\cite{shen2018reinforcewalk} proposes to solve the reward sparsity problem using off-policy learning. MultiHop~\cite{MultiHop} further extends MINERVA with reward shaping and action dropout, 
achieveing the state-of-the-art performance. These reasoning methods are intuitive and explainable. However, all of them have a weak performance in few-shot scenarios. In addition to multi-hop reasoning over KGs, there are also some multi-hop QA methods in recent years. \citet{yang2018hotpotqa} proposes a high quality dataset, which greatly promotes the development of this field. After that, many methods like CogQA \cite{CogQA} and DFGN \cite{DFGN} are also proposed.

\textbf{Meta-Learning} tries to solve the problem of ``fast adaptation on a new training task''. It has been proved to be very successful on few-shot task \cite{lake2015human, meta-translate}. Previous meta-learning models mainly focus on computer vision and imitation learning domains. In this paper, we propose a new model (Meta-KGR) using the meta-learning algorithm MAML \cite{MAML} for few-shot multi-hop reasoning. To the best of our knowledge, this work is the first research on few-shot learning for multi-hop reasoning.

\section{Model}

\begin{figure}[t]
\centering
\setlength{\abovecaptionskip}{2pt}
\setlength{\belowcaptionskip}{0pt}
\includegraphics[width=0.7\linewidth]{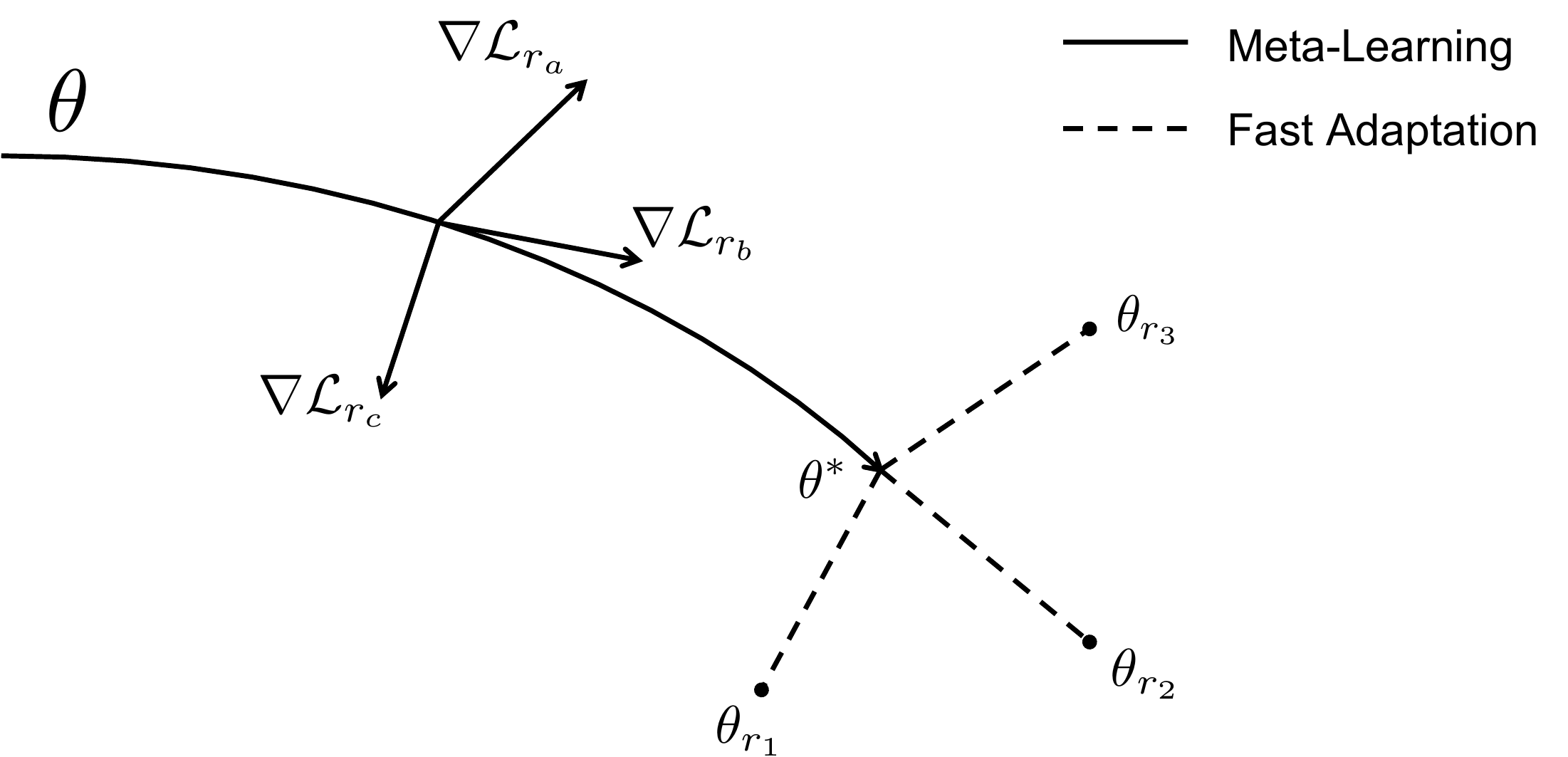}
\caption{Meta-learning.}
\label{meta}
\end{figure}

The main idea of MAML is to use a set of tasks to learn well-initialized parameters $\theta^*$ for a new task with few training examples. When applied to few-shot multi-hop reasoning, triple queries with the same relation $r$ are considered as a task $\mathcal{T}_r$. We use triples with normal relations to find well-initialized parameters $\theta^*$ and train new models on triples with few-shot relations from the found initial parameters. As shown in Figure \ref{meta}, we can easily fast adapt to new models with parameters $\theta_{r_1}$, $\theta_{r_2}$ or $\theta_{r_3}$ for few-shot relation $r_1$, $r_2$ or $r_3$. 

The learning framework of Meta-KGR can be divided into two parts: (1) relation-specific learning; (2) meta-learning. Relation-specific learning aims to learn a REINFORCE model with parameters $\theta_r$ for a specific relation $r$ (task) to search target entities and reasoning paths. Meta-learning is based on relation-specific learning and is used to learn a meta-model with parameters $\theta^*$.
We will introduce these two parts in the following sections.

\subsection{Relation-Specific Learning}

For each query relation $r_q \in \mathcal{R}$, we learn a relation-specific multi-hop reasoning agent to search for reasoning paths and target entities over $\mathcal{G}$, which is based on the on-policy RL approach proposed by~\citet{MultiHop}. 

\subsubsection{Reinforcement Learning Formulation}

The search process over $\mathcal{G}$ can be seen as a Markov Decision Process (MDP): We express $\mathcal{G}$ as a directed graph, with entities and relations as nodes and edges respectively. When given a query and its answer $(e_s, r_q, e_o)$, we expect the agent for $r_q$ can start from the source entity $e_s$ and walk through several edges to reach the target entity $e_o$. More specifically, the MDP is defined as follows.

\textbf{States} \quad The state of the $t$-th time step is defined as a tuple $s_t = (r_q, e_s, \hat{e}_t)$, where $r_q$ is the query relation, $e_s$ is the source entity and $\hat{e}_t$ is the current entity over $\mathcal{G}$. 

\textbf{Actions} \quad The action space $\mathcal{A}_t$ for a given state $s_t$ includes all outgoing edges and next entities of $\hat{e}_t$. Formally, $\mathcal{A}_t = \{(r_{t+1}, \hat{e}_{t+1})|$ $ (\hat{e}_t, r_{t+1}, \hat{e}_{t+1}) \in \mathcal{G}\}$. We add a self-loop edge to every $\mathcal{A}_t$, which is similar to a ``STOP" action.

\textbf{Transition} \quad For the state $s_t$, if the agent selects an action $(r_{t+1}, \hat{e}_{t+1}) \in \mathcal{A}_t$, the state will be changed to $s_{t+1} = (r_q, e_s, \hat{e}_{t+1})$. The transition function is defined as $\delta(s_t, \mathcal{A}_t) = (r_q, e_s, \hat{e}_{t+1})$. 
In this paper, we unroll the search over $\mathcal{G}$ up to a fixed number of time steps $T$, and finally achieve the state $s_{T} = (r_q, e_s, \hat{e}_{T})$. 

\textbf{Rewards} \quad The reward $R(s_{T}|r_q, e_s)$ will be $1$ if the agent finally stops at the right entity, i.e., $\hat{e}_{T} = e_o$, otherwise it will get an embedding-based reward $f(e_s, r_q, \hat{e}_T)$, where $f$ is a score function from knowledge graph embedding methods to measure the probability over $(e_s, r_q, \hat{e}_t)$.

\begin{algorithm}[t] 
\small
\caption{Meta-Learning for multi-hop reasoning over knowledge graphs} 
\label{algorithm-kgr}
\begin{algorithmic}[1] 
\Require 
$p(\mathcal{R})$: distribution over relations
\Require 
$\alpha, \beta$: learning rate hyperparameters
\State Randomly initialize $\theta$ for meta policy network
\While{not stop training}
\State Sample batch of relations $r \sim p(\mathcal{R})$
\ForAll{relation $r$}
\State Sample supporting set $\mathcal{D}_S$ and query set $\mathcal{D}_Q$ for task $\mathcal{T}_r$
\State Evaluate $\nabla_{\theta}L_r^{\mathcal{D}_S}(\theta)$ in Eq.~(\ref{loss_func})
\State Compute adapted parameters: $\theta'_r = \theta - \alpha \nabla_{\theta}L_r^{\mathcal{D}_S}(\theta)$
\EndFor
\State Update $\theta \leftarrow \theta - \beta \nabla_{\theta}\sum_{\mathcal{T}_r} L_r^{\mathcal{D}_Q} (\theta'_r)$
\EndWhile
\end{algorithmic} 
\end{algorithm}

\begin{table*}[t]
\small
\centering
\setlength{\belowcaptionskip}{-1pt}
    \begin{tabular}{c|ccc|ccc}
    \toprule
    \multirow{2}{*}{\textbf{Model}} & \multicolumn{3}{c|}{\textbf{FB15K-237}} & \multicolumn{3}{c}{\textbf{NELL-995}} \\
    & MRR & Hits@1 & Hits@10 & MRR & Hits@1 & Hits@10  \\
    \midrule
    NeuralLP  & 10.2  & 7.0  & 14.8   & 17.9  & 4.8 & \textbf{35.1}  \\
    NTP-$\lambda$  & 21.0  & 17.4 & 30.8   & 15.5  & 10.2 & 33.4  \\
    MINERVA  & 30.5  & 28.4  & 34.1   & 20.1  & 16.2 & 28.3  \\
    MultiHop(DistMult)  & 38.1  & 38.1  & 50.3   & 20.0  & 14.5 & 30.6  \\
    MultiHop(ConvE)  & 42.7  & 36.7  & 53.3   & 23.1  & 17.8 & 32.9  \\
    Meta-KGR(DistMult)  & 45.8  & 40.3  & 58.0   & 24.8  & \textbf{19.7} & 34.5  \\
    Meta-KGR(ConvE)  & \textbf{46.9}  & \textbf{41.2} & \textbf{58.8}   & \textbf{25.3}  & \textbf{19.7} & 34.7  \\
    \bottomrule
    \end{tabular}
    \caption{\label{table} Experimental results for link prediction. The MRR, Hits@1 and Hits@10 metrics are multiplied by 100.}
\end{table*}

\subsubsection{Policy Network}

To solve the above MDP problem, we need a model which has policy to choose actions at each state. Specifically, different from normal RL problems, we apply a policy network considering the search history over $\mathcal{G}$. Formally, after embedding all entities and relations in $\mathcal{G}$ as $\mathbf{e} \in \mathbb{R}^{d}$ and $\mathbf{r} \in \mathbb{R}^{d}$ respectively, each action $a_t = (r_{t+1}, \hat{e}_{t+1}) \in \mathcal{A}_t$ is represented as $\mathbf{a}_t = [\mathbf{r}_{t+1}; \mathbf{\hat{e}}_{t+1}]$. We use an LSTM to encode the search path,
\begin{equation}
\small
    \mathbf{h}_t = \text{LSTM} (\mathbf{h}_{t - 1}, \mathbf{a}_{t - 1}).
\end{equation}

Then, we represent the action space by stacking all actions in $\mathcal{A}_t$ as $\mathbf{A}_t \in \mathbb{R}^{|\mathcal{A}_t| \times 2d}$. The parameterized policy network is defined as,
\begin{equation}
  \small
  \pi_{\theta}(a_t|s_t) = \text{softmax} (\mathbf{A}_t (\mathbf{W}_2 \text{ReLU} (\mathbf{W}_1[\mathbf{\hat{e}}_t;\mathbf{h}_t;\mathbf{r}_q]))),
\end{equation}
where $\pi_{\theta}(a_t|s_t)$ is the probability distribution over all actions in $\mathcal{A}_t$.

\subsubsection{Loss Function}

Given a query relation $r$ and a batch of triples set $\mathcal{D}$ with relation $r$, the overall loss for this relation-specific policy network is defined as:

\begin{equation}
\label{loss_func}
\small
    \mathcal{L}_r^{\mathcal{D}}(\theta) = -\mathbb{E}_{(e_s, r, e_o) \in \mathcal{D} } \mathbb{E}_{a_1, \dots, a_{T-1} \in \pi_{\theta} } [R(s_T|e_s, r)].
\end{equation}



\subsection{Meta-Learning}



In meta-learning, we aim to learn well-initialized parameters $\theta^*$, such that small changes in the parameters will produce significant improvements on the loss function of any task \cite{MAML}.

Formally, we consider a meta policy network with parameters $\theta$. When adapting to a new task $\mathcal{T}_r$, the parameters of the model become $\theta'_r$. Following MAML, the updated parameter $\theta'_r$ is computed using one or more gradient descent updates on task $\mathcal{T}_r$. For example, assuming a single gradient step is taken with the learning rate $\alpha$
\begin{equation}
    \theta'_r = \theta - \alpha \nabla_{\theta}L_r^{\mathcal{D}_S}(\theta),
\end{equation}
where $\mathcal{D}_S$ is a supporting set randomly sampled from the triples belonging to $\mathcal{T}_r$. After the relation-specific parameters $\theta'_r$ is learned, we evaluate $\theta'_r$ on the query set $\mathcal{D}_Q$ belonging to $\mathcal{T}_r$, which is sampled like $\mathcal{D}_S$. The gradient computed from this evaluation can be used to update the meta policy network with parameters $\theta$. Usually, we will go over many tasks in a batch and update $\theta$ as follows:
\begin{equation}
    \theta = \theta - \beta \nabla_{\theta}\sum_{\mathcal{T}_r} L_r^{\mathcal{D}_Q} (\theta'_r),
\end{equation}
where $\beta$ is the meta-learning rate. We detail the meta-learning algorithm in Algorithm~\ref{algorithm-kgr}.

After previous meta-learning steps, Meta-KGR can fast adapt to a relation-specific policy network for every few-shot relation by using $\theta$ as well-initialized parameters $\theta^*$.

\begin{table}[t]
\small
\centering
\setlength{\belowcaptionskip}{-1pt}
    \begin{tabular}{lrrr}
    \toprule
    \textbf{Dataset} & \textbf{\#Ent} & \textbf{\#Rel} & \textbf{\#Triples} \\
    \midrule
    FB15k-237 (normal) & 14,448 & 200 & 268,039 \\
    FB15k-237 (few-shot) & 3,078 & 37 & 4,076 \\
    NELL-995 (normal) & 63,524 & 170 & 115,454 \\
    NELL-995 (few-shot) & 2,951 & 30 & 2,680 \\
    \bottomrule
    \end{tabular} 
    \caption{\label{table2} Statistics of datasets.}
    \vspace{-0.8em}
\end{table}

\section{Experiments}

\subsection{Datasets}

We use two typical datasets FB15K-237 \cite{FB15K-237} and NELL-995 \cite{DeepPath} for training and evaluation. Specifically, we set $K = 137$ and $K = 114$ to select few-shot relations from FB15K-237 and NELL-995 respectively. Besides, we rebuild NELL-995 to generate few-shot relations in valid and test set. Statistics are given separately for normal relations and few-shot relations in Table \ref{table2}.

\subsection{Baselines}

We compare with four multi-hop reasoning models in experiments: (1) Neural Logical Programming (NerualLP) \cite{neuralLP}; (2) NTP-$\lambda$ \cite{NTP}; (3) MINERVA \cite{MINERVA} and (4) MultiHop \cite{MultiHop}. For MultiHop and our model, we use both DistMult \cite{DistMult} and ConvE \cite{ConvE} as the reward function to create two different model variations, which are labeled in parentheses in Table \ref{table}.

\subsection{Link Prediction}

Given a query $(e_s, r_q, ?)$, link prediction for QA aims to give a ranking list of candidate entities from the KG. Following previous work \cite{ConvE, MultiHop}, we use two evaluation metrics in this task: (1) the mean reciprocal rank of all correct entities (MRR) and (2) the proportion of correct entities that rank no larger than N (Hits@N). 

Evaluation results on two datasets are shown in Table~\ref{table}. From the table, we can conclude that: (1) Our models outperform previous work in most cases, which means meta parameters learned from high-frequency relations can adapt to few-shot relations well. (2) ConvE is better than DistMult when used as the reward function in our models. This indicates that more effective knowledge graph embedding methods may provide fine-grained rewards for training multi-hop reasoning models. (3) Compared with NELL-995, FB15K-237 is denser. Our models perform well on both datasets, which demonstrates Meta-KGR can accommodate different types of KGs.

\begin{table}[t]
\small
\centering
\setlength{\belowcaptionskip}{-1pt}
    \begin{tabular}{l|cc|cc}
    \toprule
    \multirow{2}{*}{\textbf{Threshold}} & \multicolumn{2}{c|}{\textbf{MultiHop}} & \multicolumn{2}{c}{\textbf{Meta-KGR}} \\
    & MRR & Hits@1  & MRR & Hits@1  \\
    \midrule
    $K=1$  & 20.8  & 16.9  & \textbf{22.3}   & \textbf{19.3}   \\
    $K=5$  & 25.7  & 20.8  & \textbf{29.6}   & \textbf{26.6}   \\
    $K=10$  & 29.1  & 25.0  & \textbf{31.3}   & \textbf{27.2}  \\
    $K=\text{max}$  & 42.7  & 36.7  & \textbf{46.9}   & \textbf{41.2}    \\
    \bottomrule
    \end{tabular}
    \caption{\label{table3} Experimental results for robustness analysis.}
\end{table}

\subsection{Robustness Analysis}

We can use different frequency thresholds $K$ to select few-shot relations. In this section, we will study the impact of $K$ on the performance of our model. In our experiments, some triples will be removed until every few-shot relation has only $K$ triples. We do link prediction experiments on FB15K-237 and use ConvE as our reward function. The final results are shown in Table \ref{table3}. $K=\text{max}$ means we use the whole datasets in Table \ref{table2} and do not remove any triples. From Table \ref{table3} we can see our model is robust to $K$ and outperforms MultiHop in every case.

\section{Conclusion}

In this paper, we propose a meta-learning based model named Meta-KGR for multi-hop reasoning over few-shot relations of knowledge graphs. Meta-KGR uses training triples with high-frequency relations to find well-initialized parameters and fast adapt to few-shot relations. The meta information learned from high-frequency relations is helpful for few-shot relations. In experiments, our models achieve good performance on few-shot relations and outperform previous work in most cases. Some empirical analysis also demonstrates that our models are robust and generalized to different types of knowledge graphs.

\section*{Acknowledgments}
  
The work is supported by NSFC key projects (U1736204, 61533018, 61661146007), Ministry of Education and China Mobile Joint Fund (MCM20170301), and THUNUS NExT Co-Lab.

\bibliography{emnlp-ijcnlp-2019}
\bibliographystyle{acl_natbib}

\end{document}